\documentclass[11pt,a4paper]{article}
\usepackage{times,latexsym}
\usepackage{url}
\usepackage[T1]{fontenc}

\usepackage[acceptedWithA]{tacl2021v1}

\usepackage{microtype}
\usepackage{graphicx}
\usepackage{diagbox}

\usepackage{booktabs} 
\usepackage{caption}
\usepackage{subcaption}
\usepackage{multirow}
\usepackage{amssymb}  
\usepackage{amsmath}  
\usepackage{algorithm2e}  
\usepackage{amsthm}
\theoremstyle{definition}

\usepackage{xspace,mfirstuc,tabulary}

\newif\iftaclinstructions
\taclinstructionsfalse 
\iftaclinstructions
\renewcommand{\confidential}{}
\renewcommand{\anonsubtext}{(No author info supplied here, for consistency with
TACL-submission anonymization requirements)}
\newcommand{\instr}
\fi

\iftaclpubformat 

\else

\fi

\newcommand{\comment}[1]{}

\newcommand{ \our  }{\textsc{GIEA}\xspace}

\title{Getting BART to Ride the Idiomatic Train: Learning to Represent Idiomatic Expressions}

 \author{Ziheng Zeng  \and Suma Bhat \\
 Department of Electrical and Computer Engineering\\
        University of Illinois at Urbana-Champaign\\
        Champaign, IL USA\\
        {\sf \{zzeng13, spbhat2\}@illinois.edu}}

\date{}

\begin{document}
\maketitle
\begin{abstract}
  Idiomatic expressions  (IEs), characterized by their non-compositionality, are an important part of natural language. They have been a classical challenge to NLP, including pre-trained language models  that drive today's state-of-the-art. Prior work has identified deficiencies in their contextualized representation stemming from the underlying compositional paradigm of representation. In this work, we take a first-principles approach to build idiomaticity into BART using  an adapter as a lightweight
non-compositional language expert trained
on idiomatic sentences. The improved capability over baselines (e.g., BART) is seen via intrinsic and extrinsic methods, where idiom embeddings  score 0.19 points higher in homogeneity score for embedding clustering, and  up to 25\% higher  sequence accuracy on the idiom processing tasks of IE sense disambiguation and span detection.

\end{abstract}

\section{Introduction} \label{sec:introduction}
Natural language has a common yet special class of multi-word expressions (MWEs) called idiomatic expressions (IEs) that  exhibit \textit{semantic non-compositionality}, where the meaning of the expression cannot be inferred from that of its constituent words (e.g., the idiom \textit{break a leg}) \cite{DBLP:reference/nlp/BaldwinK10}.  They are commonly used  for specific communicative intents \cite{moon1998fixed, DBLP:reference/nlp/BaldwinK10} and are individually rare but collectively frequent, appearing frequently across genres \cite{moon1998fixed, haagsma2020magpie}.  They have been classically regarded as a ``pain in the neck''  to NLP systems \cite{sag2002multiword}  not only because of their non-compositionality, but also because of their contextual semantic ambiguity (used in  idiomatic or literal meaning depending on the context). Challenges posed by the presence of IEs have been identified across multiple NLU tasks even with state-of-the-art solutions, including sentiment analysis \cite{liu2017idiom, biddle2020leveraging}, paraphrase generation \cite{zhou2021idiomatic}, natural language inference \cite{chakrabarty-etal-2021-figurative}, and dialog models \cite{jhamtani-etal-2021-investigating}.

Even the the flagship NLP model GPT-3 \cite{NEURIPS2020_1457c0d6} finds idioms challenging. We tested for its idiom comprehension over 75 idioms, covering a spectrum of the most to the least frequent idioms (based on their frequency of occurrence in the BNC \cite{haagsma2020magpie}). We do this in question-answering mode where we ask GPT-3\footnote{We  used OpenAI's playground at \url{https://gpt3demo.com/}.} simple questions about idioms that can only be answered correctly if their meanings are understood in context (a sample of questions and answers are shown in Table~\ref{tab:gpt3_idiom_answers}) and noted 37 incorrect answers (50.7\% accuracy), showing its inability to comprehend many idioms.

\begin{table*}[]
\centering
\small
\caption{Questions used to probe the understanding of Idioms and the answers provided by GPT-3. }

\begin{tabular}{|p{27mm}|p{80mm}|p{35mm}|}
\hline
\textbf{Idiom} & \textbf{Question} & \textbf{Answer by GPT-3} \\ \hline
Never say die & If I have a never say die attitude, would I run the marathon injured or forfeit? & I would never run a marathon injured. \\ \hline
All at sea & If I am all at sea with my math assignment, am I making progress or am I lost? & You are making progress. \\ \hline
Every trick in the book & If I use every trick in the book to guarantee my grade, am I a magician or a determined student? & A magician. \\ \hline
Kill two birds in one stone & If I wanted to kill two birds in one stone, what kind of a workplace should I work in? & A slaughterhouse. \\ \hline
Ahead of the game & If I want to be ahead of the game, would I study early or procrastinate? & Procrastinate. \\ \hline
\end{tabular}
\label{tab:gpt3_idiom_answers}
\end{table*}

Despite the relative success of recent studies using pre-trained language models (PTLMs) for IE processing, including idiom token and type classification \cite{liu2019toward}, IE span detection \cite{zeng-bhat-2021-idiomatic, SKVORC2022107606}, and idiomatic to literal paraphrasing \cite{zhou2021idiomatic}, the emerging knowledge is that  PTLMs are unable to produce meaningful representation of IEs in task-specific contexts (e.g., for IE sense disambiguation) \cite{tayyar-madabushi-etal-2021-astitchinlanguagemodels-dataset}. 
Noting that solutions to improve IE representation in a task-agnostic manner are largely missing,   this study aims to develop targeted solutions to make language models (LMs) idiom-aware with the immediate objective of improving IE representation in large PTLMs yet without relying on astronomically large training corpora and  parameters.
In this work, we focus on improving IE representation and seek answers to the following questions about IE embeddings. \\ \textbf{Q1}: \textit{Are the current PTLMs capable of generating semantically meaningful contextualized representations for IEs?}
Answering this question, we examine the IEs embeddings  produced by a representative SOTA LM,  e.g., BART \cite{lewis-etal-2020-bart}. Specifically, we perform an intrinsic evaluation of  IE embeddings by grouping them into semantic classes and observing how they cluster. 

\begin{table*}[t]
\centering
\small
\caption{The top-3 closest idioms ranked by cosine similarity by IE embeddings generated by BART and ITI+SF+SI (our \our method). While the IE embeddings from \our are grouped by semantic meaning, BART's IE embeddings are grouped together mostly by surface-level token and/or syntactic similarity.}
\begin{tabular}{|l|l|l|}
\hline
\textbf{Idiom} & \textbf{BART} & \textbf{ ITI+SF+SI} \\ \hline
\multirow{3}{*}{in the final analysis} & in the long run & at the end of the day \\ \cline{2-3} 
 & in the works & in light of \\ \cline{2-3} 
 & in light of & all things being equal \\ \hline
\multirow{3}{*}{see red} & see the light & go spare \\ \cline{2-3} 
 & see stars & fly off the handle \\ \cline{2-3} 
 & go down like a lead balloon & do someone's head in \\ \hline
\multirow{3}{*}{quick as a flash} & flash in the pan & in the blink of an eye \\ \cline{2-3} 
 & keen as mustard & like a bat out of hell \\ \cline{2-3} 
 & thin as a rake & thick and fast \\ \hline
\end{tabular}
\label{tab:closest_idioms}
\end{table*}

Observing the low quality of IE representation, we ask, \\ \textbf{Q2}: \textit{How can we expand the capability of these LMs to produce high quality IE embeddings?} As a solution, we propose the \textbf{G}eneration of \textbf{I}diom \textbf{E}mbedding with \textbf{A}dapter (\our) approach that extends the capabilities of the current SOTA LMs by producing quality IE embeddings. 

Concretely, unlike prior work that treats each idiom as a new token \cite{hashempour-villavicencio-2020-leveraging}, \our refrains from new tokenization, to represent IEs and employs an adapter \cite{pmlr-v97-houlsby19a, pfeiffer-etal-2020-adapterhub} as a parameter-constrained learner of IE embeddings. Finally, we devise a denoising auto-encoder-style learning objective and train the network to reconstruct selective masked sentence parts. 
Our use of symbolic knowledge \cite{Yu2021DictBERTEL} of IEs to aid the learning of their embeddings results in the model needing a significantly small amount of data ($\sim$60MB) compared to that required for LM pre-training ($\sim$160GB of text for BART).  

Our main contributions are as follows. \\
\noindent (1) We demonstrate  the limited ability of SOTA PTLMs for generating semantically meaningful embeddings for IEs via a simple probing task. \\
\noindent (2) We propose a lightweight solution, \our, an adapter built over BART, to produce quality IE embeddings without altering  input sentences. \\
\noindent (3) We evaluate the resulting IE embeddings using intrinsic and extrinsic methods to show that they are meaningful in the embedding space and are task-agnostic and generalizable across different idiom processing tasks (IE sense disambiguation and IE span detection). Compared to BART, \our gains 0.19 in homogeneity score (intrinsic evaluation), performs competitively on IE sense disambiguation, and gains 25\% in sequence accuracy for IE span detection.\\
{\noindent (4) We conduct detailed analyses on the performance and limitations of \our system to provide meaningful insights and future directions\footnote{
The code for \our framework can be found at \url{https://github.com/zzeng13/GIEA}.
}.}


\section{The Inability to Represent Idiomatic Expressions} \label{sec:probelm}
Compositionality is a dominant paradigm driving the SOTA in NLP both at the tokenization and architectural levels.  The tokenization of most LMs, e.g., Byte-Pair Encoding (BPE) \cite{sennrich-etal-2016-neural} and WordPiece \cite{DBLP:journals/corr/WuSCLNMKCGMKSJL16}, assumes compositionality not only at the phrase-level but also at the word level. This suggests that the meaning of a word is deduced from that of the subword components. At the architectural level,  transformer-based LMs implicitly consider all phrases (or even words) as compositional. The self-attention mechanism in transformers considers the embedding of a word to be an attention weighted sum of the word embeddings in its context. This design leads to phrase or even sentence embeddings to be overall compositional. In addition, each IE is individually rare, compounding the difficulty for obtaining good IE representation.
This leads us to hypothesize that the inherent notion of compositionality and the rarity of IEs are a hindrance to the representation of the IEs that are inherently non-compositional. We test the validity of this hypothesis by analyzing  PTLMs' representation of IEs.

\noindent \textbf{IE embedding generation: } We first obtain the embeddings for the IEs in the MAGPIE dataset \cite{haagsma2020magpie}, { a collection of potentially idiomatic expressions (PIEs), i.e., idioms used in a literal and idiomatic sense, and the sentences in which they occur.}  Focusing on the IEs used idiomatically (thus ensuring their non-compositionality), we first retrieve all the sentences  in which they occur.  
Then, for each sentence, we extract the {BART base}  embeddings corresponding to the IE tokens in the sentence. We then apply mean pooling across the tokens and  across all the sentences in which the IE appears. In this manner we generate the embeddings for 1,480 idioms from an average of 22 sentences per idiom.
We then list IEs most similar to a set of IEs in the embedding space produced by the base BART model, computed using the cosine similarity. Table~\ref{tab:closest_idioms} shows examples of this listing including three most similar IEs (second column) to 
a sample of IEs (first column).  As noted from the examples, IEs with superficial token-level (\textit{see red} vs. \textit{see stars}) and/or syntactic-level (\textit{quick \underline{as} a flash} vs. \textit{keen \underline{as} mustard}) 
matches tend to be most similar according to BART's embeddings  without accounting for their semantic congruence. 
This  suggests that BART considers IEs mostly compositionally, an inadequate approach for representing the non-compositionality of the IEs. 

\noindent \textbf{Synonymous IE groups creation:}
To quantify the above qualitative finding, we manually assigned 129 idioms into 20 distinct meaning groups---``in summary'',
 ``anger/upset'',
 ``easy/relax'',
 ``quick'',
 ``exactly'',
 ``death'',
 ``punish/criticize'',
 ``impress'',
 ``happy'',
 ``to understand'',
 ``fail'',
 ``success'',
 ``close to'',
 ``decline/worsen'',
 ``grief/sad'',
 ``confront/deal with'',
 ``persevere'',
 ``great effort'',
 ``unimportant'',
 ``careful'',
 --- 
 averaging 6.4 idioms per group (see Table \ref{tab:example_idiom_groups} for example groups and their idioms).
 \begin{table}[]
 \caption{Example meaning groups and sampled idioms from the groups. }
 \centering
\small
\begin{tabular}{|p{12mm}|p{55mm}|}

\hline
\textbf{Group} & \textbf{Idioms} \\ \hline
Success & home and dry; bear fruit; hit the mark \\ \hline
Quick & in two shakes; full tilt; quick as a flash \\ \hline
Death & kick the bucket; drop like flies \\ \hline
Happy & on cloud nine; over the moon; ride high \\ \hline
\end{tabular}
\label{tab:example_idiom_groups}
\end{table}
{
The idiom groups must satisfy the following two requirements: (1) Any two idioms from the same group must have a similar meaning though the idioms may not necessarily be interchangeable; and (2) any two idioms from different groups must not  overlap in their meanings, i.e., the boundaries between any groups should be clear. Moreover, we selected idioms that are idiomatically monosemous (excluding  their literal interpretations) according to our dictionaries\footnote{The definitions were obtained from the Google dictionary and Wiktionary. The idiom groups will be made publicly available.}.
To group the idioms, we first created a few candidate groups based on commonly occurring idiom meanings, such as ``anger/upset'' and ``happy''. Then, for each idiom we either assigned it to an existing group or to a newly created meaning group. We only retained groups with more than three idioms and stopped the process once we had 20 groups. Using the aforementioned requirements, the validity of the groups and the idiom assignments were verified by two annotators, one with native and the other with near-native  English abilities (one of whom was not associated with this study), using an idiom dictionary as needed.  Only idiom assignments that were judged as correct by both the annotators were considered.}
 
\noindent \textbf{Clustering embeddings:} First, we generate the  embeddings  for these idioms based on their dictionary definitions using a pre-trained MPNet\footnote{The checkpoint used is  ``all-mpnet-base-v2'' from Sentence-Transformers \cite{reimers-2019-sentence-bert}.}  \cite{DBLP:conf/nips/Song0QLL20}  for sentence embeddings, referred to as \textit{definition embeddings}. As a contrast, we generate their BART IE embeddings  referred to as \textit{BART embeddings},  following the procedure discussed above. Then, we run agglomerative clustering\footnote{Implemented by Scikit-Learn \cite{scikit-learn}.} to produce 20 clusters with complete linkage using the pairwise cosine similarity between the embeddings (definition and BART embeddings separately) as the distance metric. Finally, we measure the clustering quality  using the homogeneity score as an index of the embedding quality, which is 1.0 if all the clusters contain only data points that are members of a single class. The homogeneity score for definition embeddings is $0.68$, whereas the score for BART embedding is only $0.45$. {This suggests that BART embeddings are more scattered in the embedding space with less than half of the IEs from each cluster having the same meaning.} 
 

\begin{figure*}[ht]
\centering
\includegraphics[width=0.95\textwidth]{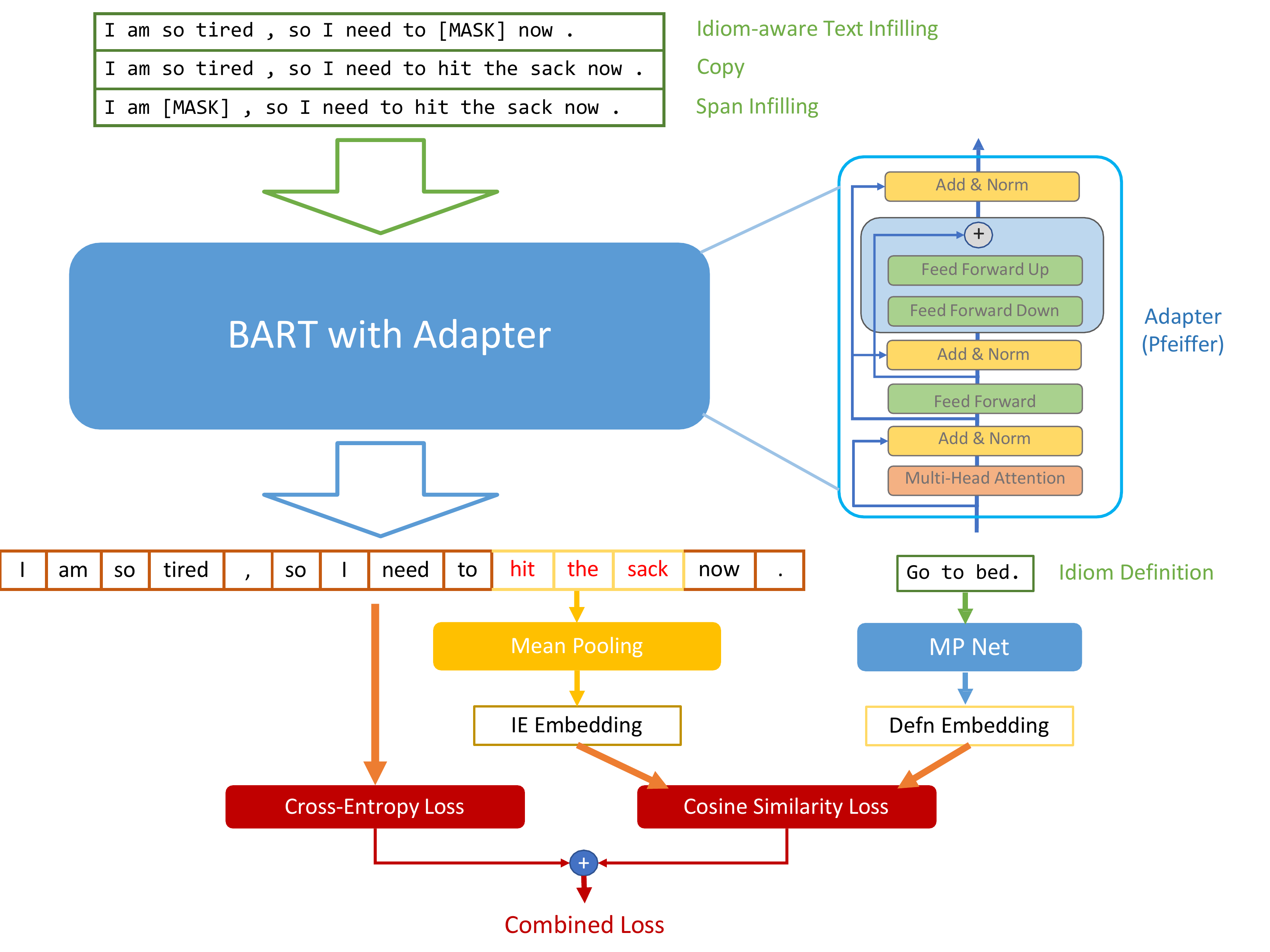}
\caption{Overview of the \our training framework.}
\label{fig:adapter_train}
\end{figure*}

\section{Learning Representation for Idiomatic Expressions}\label{sec:method}

Toward producing higher quality IE embeddings by PTLMs, we propose \our; given a set of idiomatic sentences (i.e., sentences that each contains an IE), \our freezes the base PTLM and trains an adapter that specializes in  IE representations. This is done by reconstructing idiomatic sentences that are corrupted with an idiom-aware noising function and meeting a dictionary definition-aided objective. \our's overall framework is illustrated in Figure~\ref{fig:adapter_train}. In this work, we select BART as our base PTLM. 

\noindent \textbf{Noising Function.}
Following the pre-training for BART, our training has a \textit{text corruption} stage with novel noising functions and a \textit{text reconstruction} stage. 
In the text corruption stage, we introduce three noising functions such that one permits predicting masked IEs using the context words---the \textit{idiom-aware text infilling} transformation---and the other two permit the model to use IEs to predict context words, i.e, the \textit{copy} and the \textit{span infilling} transformation.
In the idiom-aware text infilling transformation,
given a sentence containing an IE, the entire IE is replaced with a single \texttt{[MASK]} token. During training, the model is asked to reconstruct the masked IE using the context words. 
Yet, the masking of IEs alone is not sufficient for learning meaningful IE embeddings because the model sees IEs only in the decoder's input but never in the input sentences, leaving the encoder's adapter parameters unreachable by the reconstruction loss.

 The two additional noising functions,  the \textit{copy} and the \textit{span infilling} transformation, alleviate this shortcoming by allowing the model to learn to use IEs to infer the context words.  In the \textit{copy} transformation, for each sentence with its IE masked, we also supply its original, uncorrupted sentence as input and thus the model only has to copy the input sentence to the output. In the \textit{span infilling} transformation,  we mask a span of consecutive tokens \textit{excluding} the IE tokens with a single \texttt{[MASK]}, effectively asking the model to reconstruct the masked span using the IE and the remaining context. As in BART pre-training, span lengths are drawn from a Poisson distribution ($\lambda = 3$). However, our 0-length spans correspond to the original (input) sentence, identical to that of the \textit{copy} transformation. Hence, the span infilling technically subsumes the copy transformation. 
 
 Ideally, we would like the model to use an IE to predict masked context words that are directly related to the meaning of the IE. For example, as shown in Figure~\ref{fig:adapter_train}, masking the sequence  ``so tired'' helps the learning of the IE, ``hit the sack''. 
 {
 However, since the masked spans are randomly chosen, to guarantee that reconstructing the masked spans contributes to the IE meaning acquisition and inspired by prior success in prompting methods \cite{Liu2021PretrainPA},
 we inject manually created templates for span infilling (e.g., \textit{When people say hit the sack, they mean that \texttt{[MASK].}}) by connecting each IE to its dictionary  definition as a sentence. } 
  We create four such templates per idiom with variations\footnote{
The templates for a given [IE] are: \\(1) "The idiom [IE] means \texttt{[MASK]}.",\\ 
             (2) "When people say [IE] , they mean \texttt{[MASK]}.",\\ 
             (3) "[IE] is used to mean \texttt{[MASK]}.",\\
             (4) "If someone says [IE] , they mean that \texttt{[MASK]}."\\}.
             
During training, i.e., the reconstruction stage, we randomly apply the \textit{idiom-aware text infilling} transformation to 50\% of sentences, while applying the \textit{copy} or \textit{span infilling} transformation to the remaining sentences in each epoch, and the model is asked to reconstruct the uncorrupted sentences. We experiment with and analyze the use of both the \textit{copy} and \textit{span infilling}  in Section \ref{sec:results_and_analyses}.

\noindent \textbf{Similarity Forcing.} 
We leverage the  dictionary definitions of IEs to aid the learning of semantically-rich IE embeddings and supplement the small number of idiomatic sentences. 
To give an idea of the relative paucity of available idiomatic sentences,  the number of idiomatic sentences  in MAGPIE, the largest dataset for idiomatic sentences to date, is less than 30K, which is several orders of magnitude smaller than the  BART pre-training corpus. {Although collecting more sentences with IEs from other corpora is a way to directly enlarge the existing collection, isolating the truly idiomatic instances of potentially idiomatic expressions requires manual annotation, an exercise that we leave for future work.} 

Specifically, during training, we use MPNet to generate definition embeddings for each IE as before. 
{MPNet is used because it empirically outperforms BART as we will show in Section \ref{sec:results_and_analyses}.}  We also generate IE embeddings by mean pooling the BART's final layer output embeddings corresponding to the IE tokens. Note that these IE embeddings are generated from BART and the adapter being trained and thus corresponds to a non-compositional representation.
We then include the  learning objective  of increasing the cosine similarity between the IE embeddings and their corresponding definition embeddings. 
We refer to this learning objective as \textit{similarity forcing}, which is intended to  facilitate the learning of the IE embeddings by making the embedding space  be more semantically meaningful, i.e., locating IEs with similar meanings closer to each other. 

The final loss during training is the weighted sum of the cross-entropy loss from reconstruction and the cosine similarity loss from similarity forcing. In our experiments, we set the two losses to be equally weighted and leave  other weighting schemes for future explorations.

\begin{figure*}[ht]
\centering
\includegraphics[width=0.90\textwidth]{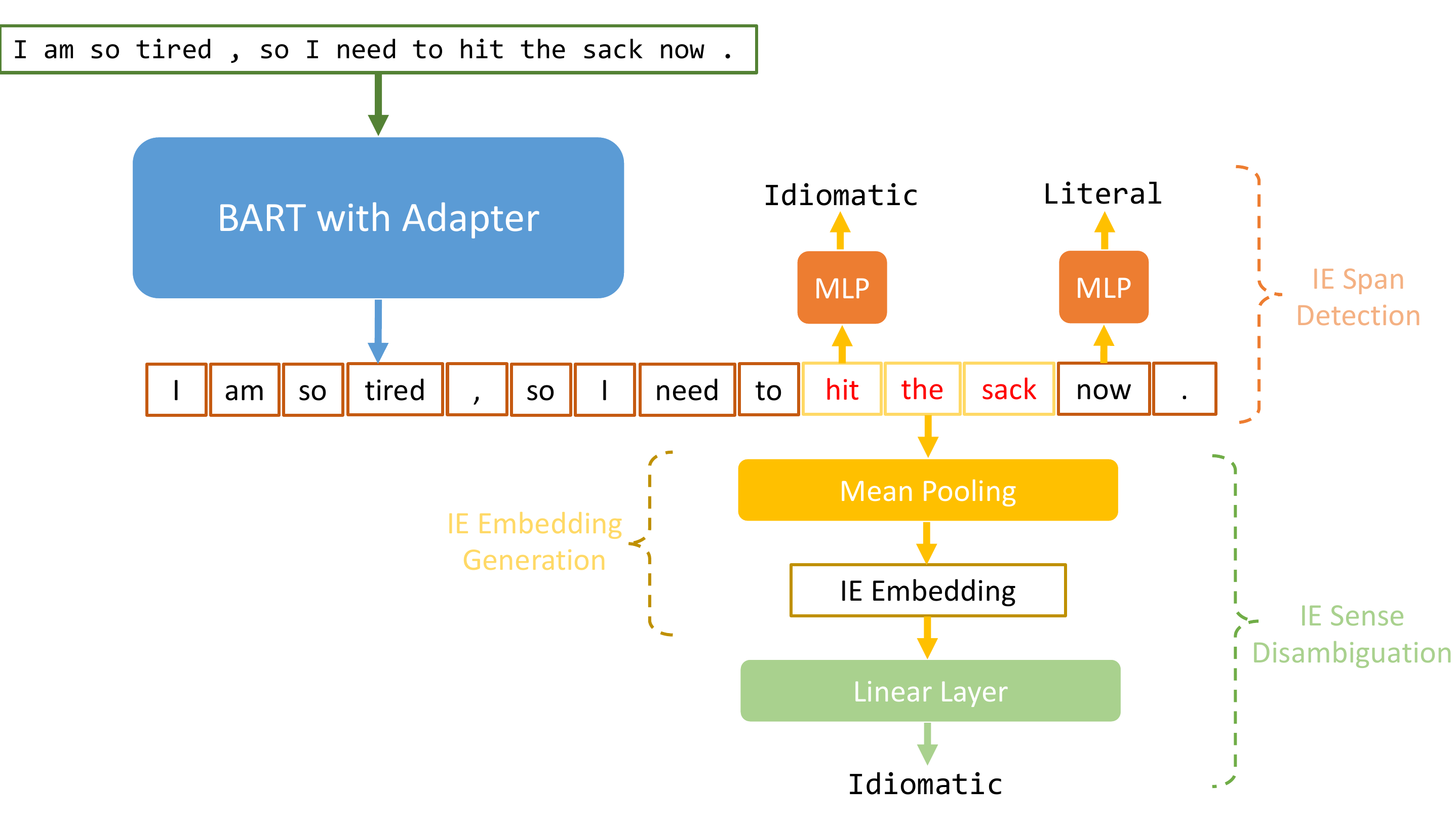}
\caption{Illustration of the intrinsic and extrinsic evaluation tasks, including the generation of IE embeddings, IE sense disambiguation, and IE span detection}
\label{fig:eval_tasks}
\end{figure*}

\noindent \textbf{Non-Compositional Language Adapter.}
Instead of fine-tuning the full model on our new learning objective, we added an adapter with the Pfeiffer architecture \cite{pfeiffer2020AdapterHub} to the base BART model for conditional generation. This is so that during training, only the parameters of the adapter are trainable while those of the underlying language model are fixed, thus making our solution lightweight. Intuitively, since the added adapter is trained with the added objective of  producing meaningful embeddings for non-compositional phrases (IEs), the adapter can be considered to be an expert in processing non-compositional language.

\section{Experiments}\label{sec:experiment}

\noindent \textbf{Datasets.} We use MAGPIE \cite{haagsma2020magpie},  a recent and the largest-to-date dataset of potentially idiomatic expressions in English,  to train \our and evaluate the baseline models. We sample a subset of the dataset by selecting idioms with a single idiomatic meaning according to our IE dictionary (referencing  Google dictionary and Wiktionary) and their corresponding sentences that are unambiguously labelled as being idiomatic (indicated by a perfect confidence score).
{The resulting collection has sentences drawn from a diverse set of genres from the British National Corpus (BNC) with 1,480 idioms with 32,693 sentences (77.4\% idiomatic) in the train set and 1,001 idioms with 4,102 (77.57\% idiomatic) sentences in the test set.}

\noindent \textbf{Evaluation Tasks.} 
The overview of the intrinsic and extrinsic evaluation tasks are illustrated in Figure~\ref{fig:eval_tasks}. The first task is an intrinsic evaluation of IE embeddings.

\noindent\textit{Embedding Clustering.}  We follow the same procedure as described in Section~\ref{sec:probelm} to perform clustering on the 20 distinct idiom groups with IE embeddings from the testing models.  
Note that we only use the sentences from the test set here to generate the IE embeddings. We use agglomerative clustering with complete linkage and pairwise embedding cosine similarity as the affinity metric.  

The following two idiom-related tasks serve as extrinsic evaluations of the IE embeddings.

\noindent\textit{IE Sense Disambiguation.} This is a common probing task used to probe if IE embeddings can differentiate the literal (compositional) from the idiomatic (non-compositional) uses of the IEs \cite{tayyar-madabushi-etal-2021-astitchinlanguagemodels-dataset, Adewumi2021PotentialIE}. 
 Many IEs can be used both figuratively or literally depending on the context. For example, the phrase ``behind closed doors'' can be interpreted literally as in \textit{The valuable items are locked behind closed doors} and can be understood figuratively as in \textit{They avoided any publicity and made all deals behind closed doors}. To account for this contextual ambiguity, these phrases are often refer to as potentially idiomatic expressions (PIEs) \cite{haagsma2020magpie}. The IE sense disambiguation task aims to classify each IE usage into idiomatic and literal class. 
To create a disambiguation classifier, we appended a single linear layer after the trained baseline embedding model. Given a sentence with a PIE and the location of the tokens belonging to the PIE, 
the baseline embedding model generates the embeddings for every token in the sentence. Then, the token embeddings corresponding to the PIE are mean pooled and fed to the linear layer to generate a binary classification. Only the linear layer is trainable when training the classifier. Given that nearly 78\% percent of IEs are used figuratively in MAGPIE test data, the majority-class baseline  predicts \textit{idiomatic} label for all instances. 

\noindent\textit{IE Span Detection.} This is a  more demanding task compared to IE sense disambiguation and studies focusing on this task are only emerging   \cite{zeng-bhat-2021-idiomatic}.  Given a sentence with a PIE, a model is expected to classify every token as \textit{idiomatic} or \textit{literal}; when the PIE is used idiomatically, the tokens from the PIE will be tagged as idiomatic; when the PIE is used literally, all its tokens will be tagged as literal. To succeed in this task, a model must identify the presence of an IE and then precisely predict its boundary. To create such a classifier, we append a two-layer MLP that reduces the number of hidden neurons by a factor of 2 after each layer and uses ReLU activation  between the layers. Only the MLP is trainable. Since the tokens are overwhelmingly literal, the majority-class baseline predicts each token to be \textit{literal}. 

Note that for both the above tasks, more powerful classifiers exist as shown in prior works \cite{liu2017representations, liu2019generalized, zeng-bhat-2021-idiomatic, SKVORC2022107606}. However, we deliberately constrain the complexity of the classifiers to linear layers (or MLPs) to ensure the performance differences reflect primarily the effect of different IE embeddings rather than that of additional modeling.  

\noindent \textbf{Evaluation Metrics.} For intrinsic evaluation, i.e., the embedding clustering task, we evaluate the performance using \textit{homogeneity score} to evaluate the clustering quality. {Given that two idioms from different groups should have distinct meanings, we also measure the \textit{mean cosine distance} between the embeddings for IEs from different groups; the larger the distance the better.} For the IE sense disambiguation task, because it is a binary classification problem, we use \textit{accuracy} and \textit{F1} score to evaluate the performance. For IE span detection, given that this is a sequence tagging task, we use three evaluation metrics, namely, \textit{sequence accuracy},\textit{ token-level recall} score and \textit{token-level accuracy}. In sequence accuracy, an instance is considered as correct if and only if all the tokens in the sequence are tagged correctly, making this the strictest metric. However, by only considering sequence accuracy, one may underestimate the performance of models that can tag most of the tokens from the positive (idiomatic) class correctly. Hence, we also consider the token-level recall and the accuracy score to complement the strict sequence accuracy metric. For token-level recall and accuracy, we compute the recall and accuracy for each predicted sequence and the final scores are averaged across all sequences.

\noindent \textbf{Baseline Models.} Due to the lack of directly related prior work, we include only the majority-class baseline, BART, and variations of \our to demonstrate the effect of different components of our method detailed below. 

\medskip
\noindent\textit{Majority-class} is a na\"{i}ve baseline that chooses the majority class for any classification problem.

\medskip
\noindent\textit{BART} is the original pre-trained BART-base model.  

\medskip
\noindent{\textit{BART-FT} is the fine-tuned full pre-trained BART-base model using dictionary definition template sentences mentioned in Section~\ref{sec:method} in addition to the MAGPIE train data with the idiom-aware text infilling and span infilling objective. 
}

\medskip
\noindent\textit{Idiom-aware Text Infilling (ITI) Model} is a baseline that trains the  adapter with only the  idiom-aware text infilling transformation. 

\medskip
\noindent\textit{Idiom-aware Text Infilling + Span Infilling Model (ITI+SI)} is a baseline that trains the adapter with both the idiom-aware text infilling and span infilling transformations.

\medskip
\noindent\textit{Idiom-aware Text Infilling + Similarity Forcing (ITI+SF)} is \our that trains the adapter with the  idiom-aware text infilling transformation and similarity forcing learning objective. 

\noindent \textbf{Our Models.} We include two competing versions of \our using different noising functions:

\noindent\textit{Idiom-aware Text Infilling + Similarity Forcing + Copy Model  (ITI+SF+Copy)} is \our that trains the adapter with the similarity forcing objective and both the idiom-aware text infilling and copy transformations. 

\medskip
\noindent\textit{Idiom-aware Text Infilling + Similarity Forcing  + Span Infilling Model (ITI+SF+SI)} is \our that trains the adapter with the similarity forcing objective and both the idiom-aware text infilling and span infilling transformations.


\noindent \textbf{Experimental Setup.} For the adapters in all baseline models, our adapter implementation  is based on \citet{pfeiffer-etal-2020-adapterhub}.  The BART-base model is implemented and maintained by Huggingface \cite{wolf-etal-2020-transformers}. The definition embeddings are generated by an MPNet hosted and maintained by the Sentence-Transformers package \cite{reimers-2019-sentence-bert}. For the adapters, we trained all baseline \our models for 220 epochs with a batch size of 16. We trained a set of IE sense disambiguation and IE span detection classifiers for each baseline model except for the majority-class baseline. For IE sense disambiguation, we trained the classifier for 55 epochs with a batch size of 32 and for IE span detection, we trained it for 100 epochs with a batch size of 16. The linear layer and the MLP in the respective classifiers were trained with a dropout rate of 0.2. For all training, we used the Adam optimizer with a learning rate of 1e-5. For all models, checkpoints with the best validation performances were used in the experiments. All the other hyperparameters were in their default values. { We only use MAGPIE's idiomatic sentences to train GIEA and the baseline models, but we use both the idiomatic and the literal sentences to train the probing models for evaluation. }

\section{Results and Analyses}\label{sec:results_and_analyses}
\begin{table}[t]
\centering
\small
\caption{Results of intrinsic evaluation via clustering. \textit{Score} is the homogeneity scores. \textit{Dist.} is the averaged cosine distance between idioms from different groups. Values are normalized (\textit{Norm.}) using BART and Definition embeddings are used as lower and upper bound. Higher values are better.}
\begin{tabular}{|l|l|l|}
\hline
\textbf{Method} & \multicolumn{1}{l|}{\textbf{Score} (Norm.)} & \multicolumn{1}{l|}{\textbf{Dist.} (Norm.)} \\ \hline
BART        & 0.4546 (0.0)      & 0.0379 (0.0)   \\ \hline
BART-FT     & 0.4659 (4.97)     & 0.0681 (14.99) \\ \hline
ITI         & 0.4597 (2.26)     & 0.0397 (0.876) \\ \hline
ITI+SI      & 0.4483 (-2.76)    & 0.0514 (6.71)  \\ \hline
ITI+SF      & 0.4357 (-8.31)    & 0.0411 (1.64)  \\ \hline
ITI+SF+Copy & 0.5906 (59.92)    & 0.1980 (79.47) \\ \hline
ITI+SF+SI   & 0.6450 (83.86)    & 0.2284 (94.54) \\ \hline
Definition  & 0.6816 (100.0)    & 0.2394 (100.0) \\ \hline
\end{tabular}
\label{tab:intrinsic_eval_clustering_perf}
\end{table}

\begin{table*}[ht]
\centering
\small
\caption{Results of IE embedding extrinsic evaluation via IE disambiguation---evaluated using F1 score (F1) and Accuracy (Acc\%), and  IE span detection---evaluated using sequence accuracy (Seq Acc\%), and token-level recall (Tkn Recall) and accuracy (Tkn Acc\%). Best performances are \textbf{boldfaced}.}
\begin{tabular}{|l|rr|rrr|}
\hline
\multirow{2}{*}{\textbf{Model}} & \multicolumn{2}{c|}{\textbf{Disambiguation}} & \multicolumn{3}{c|}{\textbf{Span Detection}} \\ \cline{2-6} 
 & \multicolumn{1}{l|}{\textbf{F1}} & \multicolumn{1}{l|}{\textbf{Acc}} & \multicolumn{1}{l|}{\textbf{Seq Acc}} & \multicolumn{1}{l|}{\textbf{Tkn Recall}} & \multicolumn{1}{l|}{\textbf{Tkn Acc}} \\ \hline
Majority Class & \multicolumn{1}{r|}{87.37} & 77.57 & \multicolumn{1}{r|}{22.43} & \multicolumn{1}{r|}{0.0} & 91.18 \\ \hline
BART & \multicolumn{1}{r|}{95.89} & 93.71 & \multicolumn{1}{r|}{50.76} & \multicolumn{1}{r|}{75.45} & 96.51 \\ \hline
BART-FT & \multicolumn{1}{r|}{96.46} & 94.49 & \multicolumn{1}{r|}{61.53} & \multicolumn{1}{r|}{84.98} & 97.24 \\ \hline
ITI & \multicolumn{1}{r|}{96.04} & 93.88 & \multicolumn{1}{r|}{55.07} & \multicolumn{1}{r|}{79.16} & 96.82 \\ \hline
ITI+SI & \multicolumn{1}{r|}{\textbf{96.53}} & \textbf{94.61} & \multicolumn{1}{r|}{60.29} & \multicolumn{1}{r|}{84.39} & 97.15 \\ \hline
ITI+SF & \multicolumn{1}{r|}{95.81} & 93.52 & \multicolumn{1}{r|}{54.97} & \multicolumn{1}{r|}{76.75} & 96.69 \\ \hline
ITI+SF+Copy & \multicolumn{1}{r|}{95.73} & 93.30 & \multicolumn{1}{r|}{\textbf{76.35}} & \multicolumn{1}{r|}{89.48} & 98.12 \\ \hline
ITI+SF+SI & \multicolumn{1}{r|}{95.73} & 93.25 & \multicolumn{1}{r|}{76.01} & \multicolumn{1}{r|}{\textbf{90.75}} & \textbf{98.17} \\ \hline
\end{tabular}
\label{tab:extrinsic_eval_perf}
\end{table*}

\noindent \textbf{Intrinsic Evaluation.}
One of the defining characteristics of a good representation is that the embedding space should be semantically meaningful, i.e., the embeddings of similar meaning IEs should be closer to each other in the embedding space via some distance metric, e.g. cosine similarity. As shown in Table \ref{tab:closest_idioms}, it is clear that after training with our ITI+SF+SI objective, the IE embeddings no longer cluster based on mere superficial similarities, instead, their meaning is the driving factor in determining their proximity in the embedding space. 
As shown in Table~\ref{tab:intrinsic_eval_clustering_perf},  the ITI+SF+SI method achieves the best homogeneity score and is significantly higher than the original BART embeddings by 0.19.  
{Also, the mean cosine distance between the embeddings for the IEs from different meaning groups is merely 0.0379 for BART, indicating the BART embeddings are inadequate in discriminating between meanings; yet, the averaged distance is 0.2284 for ITI+SF+SI, which is very close to the distance of 0.2394 by the definition embeddings.}
To provide a more direct comparison, we also normalized the  baseline performances using the BART embedding score as the lower bound and the definition embedding score as the upper bound. Comparing ITI+SF+SI and ITI+SF+Copy reveals that the more sophisticated SI noising function enabled the model to learn an embedding space that is semantically richer, as the normalized homogeneity score and cosine distance of ITI+SF+SI is higher than that of ITI+SF+Copy by 23.9 and 15.1.

\noindent \textbf{Performance on IE Sense Disambiguation.}
Though commonly used by prior work, IE sense disambiguation is a relatively simple probing task in idiom processing. As shown in Table~\ref{tab:extrinsic_eval_perf}, though ITI+SI achieves the best performance numerically, all methods compared achieve competitive performances w.r.t F1 and accuracy. This shows that BART embeddings already capture the idiosyncratic properties of IEs, in line with the findings from recent papers \cite{tayyar-madabushi-etal-2021-astitchinlanguagemodels-dataset, Adewumi2021PotentialIE}.
However, we believe that one cannot judge the quality of IE embeddings via this task alone, because IE senses can be distinguished correctly without the semantic knowledge of IEs. 
{As an evidence, under the same setting, we trained another disambiguation classifier with BART but replaced all the IEs from the sentences with single mask tokens for the classifier to make predictions based on just the embeddings of the mask tokens, thus removing all possible IE-related semantic information. We found that such a classifier still performs with an 86\% accuracy, operating only on non-IE contextual information. 
So, IE comprehension ability and IE embedding quality cannot be fully assessed by probing the IE sense disambiguation ability, suggesting that the intrinsic embedding quality and performances on more difficult IE processing tasks must also be considered. 
}

\noindent \textbf{Performance on IE Span Detection.}
IE span detection is more difficult than IE sense disambiguation as it requires detecting the presence of IEs and precisely identifying their locations. The performance in this task showcases the superiority of our IE embedding methods. ITI+SF+Copy achieves the best performance that is 25.6 points higher than BART in sequence accuracy, our strictest metric. For token-level recall and token-level accuracy,  ITI+SF+SI achieves the best performance with a 15-point gain in recall and 1.66 higher in accuracy than BART. The gain in token accuracy is small because the tokens are overwhelmingly literal; the majority-class baseline already achieves a 91\% accuracy. The fact that ITI+SF+SI has better token-level performance than ITI+SF+Copy signifies that though ITI+SF+SI detects the span less precisely, it recovers the tokens from within IEs better than ITI+SF+Copy does.

\noindent \textbf{Effect of Copy and Span Infilling.}
We next examine the usefulness of the copy transformation and span infilling transformation in the noising function. Without copy and span infilling, the  ITI+SF suffers in both intrinsic and extrinsic evaluation. 
For embedding clustering, the homogeneity score of ITI+SF is lower than ITI+SF+SI by 0.15 and lower than ITI+SF+Copy by 0.21, performing even slightly worse than the original BART's embeddings.
For IE span detection, ITI+SF's sequence accuracy  is lower than that of ITI+SF+SI  and ITI+SF+Copy by 21.0\% and  21.4\% respectively. Notably, without copy and span infilling transformation, ITI+SF performs barely better than BART, gaining only 4.2\% in sequence accuracy. 
To a lesser degree, ITI+SI also demonstrates the usefulness of the span infilling transformation when compared with ITI, gaining 5.22\% in sequence accuracy.  
Thus, copy or span infilling transformation is necessary and beneficial during the training of the embedding model. 
Moreover, even though  ITI+SF+Copy and ITI+SF+SI performs competitively on the extrinsic evaluation tasks, ITI+SF+SI outperforms ITI+SF+Copy in the intrinsic evaluation task by a meaningful margin demonstrating  ITI+SF+SI's superiority over ITI+SF+Copy.

\begin{table*}[t]
\caption{Alternative models' evaluation performances with different LM base models and sentence embedding models (Sent Emb). All models are trained with the same ITI+SF+SI objective.}
\centering
\small
\begin{tabular}{|l|l|r|rr|rrr|}
\hline
\multicolumn{1}{|c|}{\multirow{2}{*}{\textbf{Base Model}}} & \multicolumn{1}{c|}{\multirow{2}{*}{\textbf{Sent Emb}}} & \multicolumn{1}{c|}{\textbf{Clustering}} & \multicolumn{2}{c|}{\textbf{Disambiguation}} & \multicolumn{3}{c|}{\textbf{Span Detection}} \\ \cline{3-8} 
\multicolumn{1}{|c|}{} & \multicolumn{1}{c|}{} & \multicolumn{1}{l|}{\textbf{Homogeneity}} & \multicolumn{1}{l|}{\textbf{F1}} & \multicolumn{1}{l|}{\textbf{Acc}} & \multicolumn{1}{l|}{\textbf{Seq Acc}} & \multicolumn{1}{l|}{\textbf{Tkn Recall}} & \multicolumn{1}{l|}{\textbf{Tkn Acc}} \\ \hline
BART & MPNet & 0.6450 & \multicolumn{1}{r|}{95.73} & 93.25 & \multicolumn{1}{r|}{76.01} & \multicolumn{1}{r|}{90.75} & 98.17 \\ \hline
BART & BART & 0.4671 & \multicolumn{1}{r|}{95.75} & 93.29 & \multicolumn{1}{r|}{74.55} & \multicolumn{1}{r|}{88.66} & 98.02 \\ \hline
BERT & MPNet & 0.4879 & \multicolumn{1}{r|}{91.42} & 86.36 & \multicolumn{1}{r|}{56.05} & \multicolumn{1}{r|}{78.19} & 97.34 \\ \hline
\end{tabular}

\label{tab:alt_model_ext_performances}
\end{table*}

\noindent \textbf{Effect of Similarity Forcing.}
By comparing ITI+SF and ITI or ITI+SF+SI and ITI+SI, we examine the effect of similarity forcing. 
While ITI+SF performs similarly or even slightly worse than ITI on evaluation tasks, the performance gain of ITI+SF+SI over ITI+SI is noteworthy, e.g., it gains 15.8\% in sequence accuracy for IE span detection and 0.20 points in homogeneity score for embedding clustering. Considering  the effect of copy and span infilling noising function, we see that  ITI+SF+SI shows better performance than either ITI+SI or ITI+SF. This leads us to infer that similarity forcing is only useful when combined with the copy and span infilling transformation. {In addition, we also compare the performance between ITI+SF+SI and BART-FT to demonstrate the usefulness of similarity forcing. BART-FT is a BART model fine-tuned on the same training data as ITI+SI. Though BART-FT has significantly more trainable parameters and the same access to external knowledge from the IE definition template sentences during training, BART-FT under-performs ITI+SF+SI by 14.48 points in sequence accuracy for span detection and 0.18 points in homogeneity score for embedding clustering. 
Therefore, we conclude that using similarity forcing in combination with \textit{copy-} or \textit{span infilling} transformation can boost the performance by a significant margin.}

\noindent \textbf{MPNet vs.\ BART for Definition Embedding.}
{
Though the MPNet's definition embeddings and BART's IE embeddings are in different spaces, we believe  minimizing the cosine similarity between them to improve IE embeddings' semantic meanings is a valid exercise because (1) the idiomatic meanings of IEs and the meaning of their component words are not related; hence relating their idiomatic meanings to the definition meanings from MPNet's space will not affect the embeddings of the original words; and (2) prior research suggests that minimizing cosine similarity can even help relate the meanings between image embeddings and natural language embeddings (clearly not in the same embedding space) \cite{Radford2021LearningTV}, hence the space difference between MPNet and BART should not present a problem. Moreover, using MPNet for the definition embedding results in an overall better empirical performance because MPNet produces higher-quality sentence embeddings than BART. We experimented training the ITI+SF+SI model but replaced the MPNet's definition embeddings with that from BART. Comparing the results of the resulting model with  those of  ITI+SF+SI with MPNet embeddings, shown in the second row of Table~\ref{tab:alt_model_ext_performances}, we see the resulting model achieves competitive performance for disambiguation but inferior performances in both span detection and embedding clustering with a sequence accuracy that is lower by 1.46\% and a homogeneity score that is lower by 0.18. In fact, even the \textit{definition embedding}, when generated by BART, only obtains a homogeneity score of 0.55 (not shown in tables) which is even lower than the ITI+SF+SI by around 0.10. This justifies our use of MPNet for definition embeddings. 
}

\noindent \textbf{Effect of Base Language Models.}
{
In our case, encoder-decoder LMs, e.g., BART, are more suitable than an encoder-only LMs, e.g., BERT, because the decoder allows the use of the idiom-aware text infilling objective that asks the model to reconstruct the entire idiom from a \textit{single} mask token. To empirically demonstrate the benefit, we trained an ITI+SF+SI model with BERT as the base LM and modified the idiom-aware text infilling objective by using one mask token per idiom token. As shown in the third row of Table~\ref{tab:alt_model_ext_performances}, the BERT-based model under-performs its BART-based counterpart in all evaluation tasks by large margins.
}


\noindent \textbf{Error Analysis on IE Embeddings.}
{
Here, we further examine the quality of the definition embeddings and  ITI+SF+SI's IE embeddings  (named \textit{GIEA embeddings}).  We compute \textit{precision at k} (P@k) score for each idiom from the 129 idioms in the 20 meaning groups as follows. Given the embedding for an IE, $E$, we first find the $k=3$ closest IEs using pairwise cosine similarity and $n$, the number of $k$ closest IEs that are from the same group as $E$; then, P@3 is computed as $n/k$.
The mean score for definition embeddings is $0.64$. Meanwhile, the mean score for \our embeddings is $0.52$, i.e., each IE has about half of the $3$-closest IEs from the same group. We found a large disparity among the groups with respect to the mean score for each meaning group. While most groups have a mean score around 0.5, groups such as `anger/upset', `quick', and `success' have scores higher than 0.6, and those of others, such as `punish/criticize', `decline/worsen', `persevere' are lower than 0.2. 
Also, we found that the per group P@3 scores of the definition embedding are positively correlated with those of \our embedding  with a Pearson correlation coefficient of 0.76. 
Based on these observations, we infer that the difficulty of learning IE  meanings depends on the specific meaning group and the quality of the definition embedding directly affects the learned \our embedding. Improving the definition embeddings through better sentence embedding methods (e.g., by training specifically on dictionary definitions) may further improve the performance of our method. We also leave the important aspect assessing the quality of original compositional embeddings after learning IE embeddings to a follow-up study.
}

\noindent \textbf{Error Analysis on Extrinsic Evaluation Tasks.}
{
Here, we analyze the error of the best performing ITI+SF+SI model on the tasks of span detection and disambiguation. For span detection, we sampled 300 incorrect instances with imperfect sequence accuracies (30.5\% of all incorrect samples) and categorized them into the six error types defined in \citet{zeng-bhat-2021-idiomatic}. Among the sampled errors, we found that 3.7\% were attributable to identifying one of the IEs when multiple IEs are present, 57\% to detecting only a portion of the idiom span, 1\% to identifying figurative expressions other than the ground truth idiom, 25\% to identifying a PIE 
as idiomatic when actually used in the literal sense, 8.3\% for failing to recognize the presence of an idiom, and another 5\% for returning random tokens that are not meaningful nor part of any PIEs, i.e., over 60\% of the errors  were in the detection of figurative tokens. In fact, over 40.8\% of test idioms had their spans precisely tagged in all of their test instances. For disambiguation, over 82.8\% of the test PIEs were classified with 100\% accuracy and only less than 6\% of the test PIEs had an accuracy less than 50\%. For both disambiguation and span detection, the per-idiom accuracies were weakly correlated with the number of training instances per idiom (Pearson correlation coefficient of -3.84e-4 for disambiguation and 0.26 for span detection){, suggesting that the performance discrepancy among idiom types is caused by factors other than their frequency in the train set.}
Future  studies should consider the characteristics of the hard-to-learn idioms to improve the embeddings of the under-performing idioms. 
}

\noindent \textbf{Limitations.}
{
An obvious limitation of \our is that it cannot generalize its representation ability to idioms unseen  during training. From the results in Section~\ref{sec:probelm} and Section~\ref{sec:results_and_analyses}, it is evident that the meanings of IEs cannot be learned from general corpora alone (even when  there is a collection of sentences with IEs), rather, external knowledge, e.g., IE definitions, are a fundamental to providing the strong supervising signal (i.e., similarity forcing loss) needed for training. Taking this into consideration, we believe that it is impractical to generalize the representation ability to the unseen idioms because (1) intuitively, each IE has a unique origin, metaphorical linkage, and interpretation, so, the \textit{meaning} of IEs have to be learned on a case-by-case basis; and (2) from our error analysis, even with the same training data and objective, the learning difficulty is highly  idiom dependent, a point that is also corroborated by \citet{10.3389/frai.2022.813967}. 
Therefore, we do not currently see a practical way to generalize \our to idioms that are unseen. 
However, we argue that this does not hinder the utility of \our, since our training data, MAGPIE  already contains idiomatic sentences for idioms (and metaphors) that occur in sources such as the Oxford Dictionary of English Idioms \cite{nla.cat-vn4806420}
and Wiktionary. Thus, we expect the \our to  cover most frequently used idioms. Besides, even though expanding an IE lexicon to include new idioms may be easy, gathering idiomatic sentences for those new idioms requires human input. So, an important future study is to consider methods that generalize \our to idioms with known identities  but with limited or no idiomatic sentences.
}

\section{Related Work} \label{sec:related_work}
\noindent \textbf{IE Processing Tasks.} 
Classically, two main idiom-related processing tasks, namely, \textit{idiom type classification} and \textit{idiom token classification} have been studied \cite{cook2008vnc, liu2019generalized, liu2019toward}. Idiom type classification aims to decide if a set of MWEs can be used as IEs without considering additional context 
\cite{westerstaahl2002compositionality, fazly2006automatically, tabossi2008processing, tabossi2009idioms, shutova2010metaphor, reddy2011empirical, cordeiro2016predicting}. Idiom token classification  determines if a given PIE is used in a literal or figurative sense in a sentence and solutions include those that mostly assume the knowledge of the location and/or identify of the PIEs \cite{fazly2009unsupervised,feldman2013automatic, DBLP:conf/simbig/PengF16a, salton2016idiom, taslimipoor2018identification, peng2018classifying, liu2019generalized}, build per-idiom classifiers \cite{rajani2014using, liu2017representations}, extract embeddings based on PIE positions \cite{liu2019generalized}, or focus on only PIEs with specific syntactic structures \cite{taslimipoor2018identification}. Due to the impracticality of acquiring these prior knowledge in real-world applications, most recent works \cite {zeng-bhat-2021-idiomatic, SKVORC2022107606} study the \textit{idiomatic expression identification problem}, jointly the detecting and localizing a PIE without requiring PIE identity or position. {This problem is related to the MWE identification task in STREUSLE \cite{schneider-smith-2015-corpus} but with a focus on expressions with semantic idiomaticity}. In-line with prior art, we use the IE token classification and IE identification, dubbed as \textit{IE sense disambiguation} and \textit{IE span detection}, as the extrinsic evaluation tasks to our IE embeddings.

\noindent \textbf{Impact of IE Presence.} Since \citet{sag2002multiword}'s study on the impact of MWE, not only have studies identified the influence of  IEs across various NLP applications \cite{salton2014empirical, fadaee2018examining,ganitkevitch2013ppdb,liu2017idiom, biddle2020leveraging},  recent efforts have also sought ways to mitigate them  \cite{jhamtani-etal-2021-investigating,chakrabarty-etal-2021-figurative}. 
However, the techniques used either simply enlarge the training data by including idiomatic sentences or paraphrase idiomatic sentences into equivalent literal sentences, 
completely ignoring the fundamental issue of IE representation. Other works \cite{tayyar-madabushi-etal-2021-astitchinlanguagemodels-dataset} have probed  how idiomaticity is handled in PTLMs but offer no solution to improve their representation. 
 Efforts to improve IE span detection or IE sense disambiguation include transforming the original representations from pre-trained LMs by incorporating static word embeddings alone \cite{liu2017representations}, with additional syntactic information  \cite{zeng-bhat-2021-idiomatic}, utilizing contrastive loss to make   literal and figurative speech embeddings more distinctive \cite{lin-etal-2021-cate}, treating IEs as new tokens during training \cite{hashempour-villavicencio-2020-leveraging}, or  combining representations from multiple pre-trained LMs \cite{SKVORC2022107606}. Taking a different approach in this work, instead of creating task-specific representations or altering tokenization at the input, we first train an LM that produces better IE embeddings in general and then show their benefit 
 in the idiom processing tasks. In principle, our trained \our can be plugged into the prior works for idiom processing tasks, replacing their embedding models and improving their performances, an aspect we leave to future explorations.

\noindent \textbf{Adapter.} Originally developed for computer vision applications  \cite{NIPS2017_e7b24b11, 8578945}, adapters are new modules of simple projection layers added between the trained transformer layers, used in NLP as a parameter-efficient and fast fine-tuning method to adapt pre-trained LMs to new tasks or domains \cite{pmlr-v97-houlsby19a, bapna-firat-2019-simple}. 
Recently, adapters have shown effectiveness in multi-task and multi-lingual  transfer learning as well \cite{pfeiffer-etal-2020-mad, ansell-etal-2021-mad-g}. In this work, we utilize an adapter as a lightweight non-compositional language expert that is trained on idiomatic sentences and thus can expand upon the base LM to generate semantically meaningful IE embeddings. The compact Pfeiffer adapter architecture \cite{pfeiffer-etal-2020-adapterhub} is used in \our. 

\noindent \textbf{(Non-)Compositional Phrase Embedding.} 
The core idea for works on non-compositional phrase embeddings is to avoid treating phrases as purely compositional (by aggregating word embeddings) or non-compositional (treating phrases as single units), but consider both  aspects. The approaches have adaptive
weights and consider different compositions within a phrase  \cite{ijcai2018-576,hashimoto-tsuruoka-2016-adaptive, li2018phrase} or  utilize hypernymy information and represent phrases in special embedding spaces  \cite{jana-etal-2019-compositionality}. 
Although related, these embedding methods cannot produce the contextualized phrase embeddings  as transformer-based models do, nor can be combined with PTLMs  to aid  downstream  tasks. 

\noindent \textbf{Embedding Evaluation.}  
The evaluation of word and  phrase embeddings \cite{hashimoto-tsuruoka-2016-adaptive, jana-etal-2019-compositionality} is typically via \textit{intrinsic} methods (e.g., similarity and analogy) and \textit{extrinsic} methods, e.g., downstream NLP tasks 
 \cite{schnabel-etal-2015-evaluation, ghannay-etal-2016-word, ijcai2018-796, wang_wang_chen_wang_kuo_2019}. 
A popular alternative evaluation method is \textit{probing}, where a simple diagnostic classifier is trained to extract information from frozen embeddings 
and  determine the extent to which desired linguistic properties are encoded in the representations \cite{DBLP:journals/corr/AdiKBLG16,warstadt-etal-2019-investigating, alt-etal-2020-probing, ravichander-etal-2021-probing}. Our intrinsic and extrinsic evaluation of embeddings follow these prior works. 

\section{Conclusion and Future Work}\label{sec:conclusion}
In this work, we first demonstrate current BART's inability produce semantically meaningful representations for idioms, then, we propose \our, that uses  a lightweight adapter, a set of denoising auto-encoder-style learning objectives,  and a similarity forcing objective to produce quality IE embeddings without altering the input tokenization. Through both intrinsic evaluation of embedding quality and extrinsic evaluation on their usefulness on idiom-processing tasks, we find that \our 
greatly improves upon embedding quality and usefulness compared to the original pre-trained BART's embeddings.

{
Future work should explore means to improve embedding quality for hard-to-learn idioms based on observed performance, IEs other than idioms (e.g., phrasal verbs), and  the use of  \our with other SOTA idiom processing models. 
Lastly, applying idiom-aware PTLMs to downstream applications that require the IE comprehension, such as dialog modeling and machine translation would be fruitful pursuits. 
}

\bibliography{reference}

\begin{thebibliography}{70}
\expandafter\ifx\csname natexlab\endcsname\relax\def\natexlab#1{#1}\fi

\bibitem[{Adewumi et~al.(2021)Adewumi, Javed, Vadoodi, Tripathy, Nikolaidou,
  Liwicki, and Liwicki}]{Adewumi2021PotentialIE}
Tosin~P. Adewumi, Saleha Javed, Roshanak Vadoodi, Aparajita Tripathy,
  Konstantina Nikolaidou, Foteini~Simistira Liwicki, and Marcus Liwicki. 2021.
\newblock Potential idiomatic expression (pie)-english: Corpus for classes of
  idioms.
\newblock \emph{ArXiv}, abs/2105.03280.

\bibitem[{Adi et~al.(2016)Adi, Kermany, Belinkov, Lavi, and
  Goldberg}]{DBLP:journals/corr/AdiKBLG16}
Yossi Adi, Einat Kermany, Yonatan Belinkov, Ofer Lavi, and Yoav Goldberg. 2016.
\newblock \href {http://arxiv.org/abs/1608.04207} {Fine-grained analysis of
  sentence embeddings using auxiliary prediction tasks}.
\newblock \emph{CoRR}, abs/1608.04207.

\bibitem[{Alt et~al.(2020)Alt, Gabryszak, and Hennig}]{alt-etal-2020-probing}
Christoph Alt, Aleksandra Gabryszak, and Leonhard Hennig. 2020.
\newblock \href {https://doi.org/10.18653/v1/2020.acl-main.140} {Probing
  linguistic features of sentence-level representations in neural relation
  extraction}.
\newblock In \emph{Proceedings of the 58th Annual Meeting of the Association
  for Computational Linguistics}, pages 1534--1545, Online. Association for
  Computational Linguistics.

\bibitem[{Ansell et~al.(2021)Ansell, Ponti, Pfeiffer, Ruder, Glava{\v{s}},
  Vuli{\'c}, and Korhonen}]{ansell-etal-2021-mad-g}
Alan Ansell, Edoardo~Maria Ponti, Jonas Pfeiffer, Sebastian Ruder, Goran
  Glava{\v{s}}, Ivan Vuli{\'c}, and Anna Korhonen. 2021.
\newblock \href {https://doi.org/10.18653/v1/2021.findings-emnlp.410}
  {{MAD}-{G}: {M}ultilingual adapter generation for efficient cross-lingual
  transfer}.
\newblock In \emph{Findings of the Association for Computational Linguistics:
  EMNLP 2021}, pages 4762--4781, Punta Cana, Dominican Republic. Association
  for Computational Linguistics.

\bibitem[{Ayto and Press.(2009)}]{nla.cat-vn4806420}
John. Ayto and Oxford~University Press. 2009.
\newblock \emph{Oxford dictionary of English idioms / [edited] by John Ayto},
  3rd ed. edition.
\newblock Oxford University Press [Oxford].

\bibitem[{Baldwin and Kim(2010)}]{DBLP:reference/nlp/BaldwinK10}
Timothy Baldwin and Su~Nam Kim. 2010.
\newblock \href {http://www.crcnetbase.com/doi/abs/10.1201/9781420085938-c12}
  {Multiword expressions}.
\newblock In Nitin Indurkhya and Fred~J. Damerau, editors, \emph{Handbook of
  Natural Language Processing, Second Edition}, pages 267--292. Chapman and
  Hall/CRC.

\bibitem[{Bapna and Firat(2019)}]{bapna-firat-2019-simple}
Ankur Bapna and Orhan Firat. 2019.
\newblock \href {https://doi.org/10.18653/v1/D19-1165} {Simple, scalable
  adaptation for neural machine translation}.
\newblock In \emph{Proceedings of the 2019 Conference on Empirical Methods in
  Natural Language Processing and the 9th International Joint Conference on
  Natural Language Processing (EMNLP-IJCNLP)}, pages 1538--1548, Hong Kong,
  China. Association for Computational Linguistics.

\bibitem[{Biddle et~al.(2020)Biddle, Joshi, Liu, Paris, and
  Xu}]{biddle2020leveraging}
Rhys Biddle, Aditya Joshi, Shaowu Liu, Cecile Paris, and Guandong Xu. 2020.
\newblock Leveraging sentiment distributions to distinguish figurative from
  literal health reports on {Twitter}.
\newblock In \emph{Proceedings of The Web Conference 2020}, pages 1217--1227.

\bibitem[{Brown et~al.(2020)Brown, Mann, Ryder, Subbiah, Kaplan, Dhariwal,
  Neelakantan, Shyam, Sastry, Askell, Agarwal, Herbert-Voss, Krueger, Henighan,
  Child, Ramesh, Ziegler, Wu, Winter, Hesse, Chen, Sigler, Litwin, Gray, Chess,
  Clark, Berner, McCandlish, Radford, Sutskever, and
  Amodei}]{NEURIPS2020_1457c0d6}
Tom Brown, Benjamin Mann, Nick Ryder, Melanie Subbiah, Jared~D Kaplan, Prafulla
  Dhariwal, Arvind Neelakantan, Pranav Shyam, Girish Sastry, Amanda Askell,
  Sandhini Agarwal, Ariel Herbert-Voss, Gretchen Krueger, Tom Henighan, Rewon
  Child, Aditya Ramesh, Daniel Ziegler, Jeffrey Wu, Clemens Winter, Chris
  Hesse, Mark Chen, Eric Sigler, Mateusz Litwin, Scott Gray, Benjamin Chess,
  Jack Clark, Christopher Berner, Sam McCandlish, Alec Radford, Ilya Sutskever,
  and Dario Amodei. 2020.
\newblock \href
  {https://proceedings.neurips.cc/paper/2020/file/1457c0d6bfcb4967418bfb8ac142f64a-Paper.pdf}
  {Language models are few-shot learners}.
\newblock In \emph{Advances in Neural Information Processing Systems},
  volume~33, pages 1877--1901. Curran Associates, Inc.

\bibitem[{Chakrabarty et~al.(2021)Chakrabarty, Ghosh, Poliak, and
  Muresan}]{chakrabarty-etal-2021-figurative}
Tuhin Chakrabarty, Debanjan Ghosh, Adam Poliak, and Smaranda Muresan. 2021.
\newblock \href {https://doi.org/10.18653/v1/2021.findings-acl.297} {Figurative
  language in recognizing textual entailment}.
\newblock In \emph{Findings of the Association for Computational Linguistics:
  ACL-IJCNLP 2021}, pages 3354--3361, Online. Association for Computational
  Linguistics.

\bibitem[{Cook et~al.(2008)Cook, Fazly, and Stevenson}]{cook2008vnc}
Paul Cook, Afsaneh Fazly, and Suzanne Stevenson. 2008.
\newblock The {VNC}-tokens dataset.
\newblock In \emph{Proceedings of the LREC Workshop Towards a Shared Task for
  Multiword Expressions (MWE 2008)}, pages 19--22.

\bibitem[{Cordeiro et~al.(2016)Cordeiro, Ramisch, Idiart, and
  Villavicencio}]{cordeiro2016predicting}
Silvio Cordeiro, Carlos Ramisch, Marco Idiart, and Aline Villavicencio. 2016.
\newblock Predicting the compositionality of nominal compounds: Giving word
  embeddings a hard time.
\newblock In \emph{Proceedings of the 54th Annual Meeting of the Association
  for Computational Linguistics (Volume 1: Long Papers)}, pages 1986--1997.

\bibitem[{Fadaee et~al.(2018)Fadaee, Bisazza, and Monz}]{fadaee2018examining}
Marzieh Fadaee, Arianna Bisazza, and Christof Monz. 2018.
\newblock \href {https://aclanthology.org/L18-1148} {Examining the tip of the
  iceberg: A data set for idiom translation}.
\newblock In \emph{Proceedings of the Eleventh International Conference on
  Language Resources and Evaluation ({LREC} 2018)}, Miyazaki, Japan. European
  Language Resources Association (ELRA).

\bibitem[{Fazly et~al.(2009)Fazly, Cook, and Stevenson}]{fazly2009unsupervised}
Afsaneh Fazly, Paul Cook, and Suzanne Stevenson. 2009.
\newblock Unsupervised type and token identification of idiomatic expressions.
\newblock \emph{Computational Linguistics}, 35(1):61--103.

\bibitem[{Fazly and Stevenson(2006)}]{fazly2006automatically}
Afsaneh Fazly and Suzanne Stevenson. 2006.
\newblock Automatically constructing a lexicon of verb phrase idiomatic
  combinations.
\newblock In \emph{11th Conference of the European Chapter of the Association
  for Computational Linguistics}.

\bibitem[{Feldman and Peng(2013)}]{feldman2013automatic}
Anna Feldman and Jing Peng. 2013.
\newblock Automatic detection of idiomatic clauses.
\newblock In \emph{International Conference on Intelligent Text Processing and
  Computational Linguistics}, pages 435--446. Springer.

\bibitem[{Ganitkevitch et~al.(2013)Ganitkevitch, Van~Durme, and
  Callison-Burch}]{ganitkevitch2013ppdb}
Juri Ganitkevitch, Benjamin Van~Durme, and Chris Callison-Burch. 2013.
\newblock {PPDB}: The paraphrase database.
\newblock In \emph{Proceedings of the 2013 Conference of the North American
  Chapter of the Association for Computational Linguistics: Human Language
  Technologies}, pages 758--764.

\bibitem[{Ghannay et~al.(2016)Ghannay, Favre, Est{\`e}ve, and
  Camelin}]{ghannay-etal-2016-word}
Sahar Ghannay, Benoit Favre, Yannick Est{\`e}ve, and Nathalie Camelin. 2016.
\newblock \href {https://aclanthology.org/L16-1046} {Word embedding evaluation
  and combination}.
\newblock In \emph{Proceedings of the Tenth International Conference on
  Language Resources and Evaluation ({LREC}'16)}, pages 300--305,
  Portoro{\v{z}}, Slovenia. European Language Resources Association (ELRA).

\bibitem[{Haagsma et~al.(2020)Haagsma, Bos, and Nissim}]{haagsma2020magpie}
Hessel Haagsma, Johan Bos, and Malvina Nissim. 2020.
\newblock {MAGPIE}: A large corpus of potentially idiomatic expressions.
\newblock In \emph{Proceedings of The 12th Language Resources and Evaluation
  Conference}, pages 279--287.

\bibitem[{Hashempour and
  Villavicencio(2020)}]{hashempour-villavicencio-2020-leveraging}
Reyhaneh Hashempour and Aline Villavicencio. 2020.
\newblock \href {https://aclanthology.org/2020.cogalex-1.9} {Leveraging
  contextual embeddings and idiom principle for detecting idiomaticity in
  potentially idiomatic expressions}.
\newblock In \emph{Proceedings of the Workshop on the Cognitive Aspects of the
  Lexicon}, pages 72--80, Online. Association for Computational Linguistics.

\bibitem[{Hashimoto and Tsuruoka(2016)}]{hashimoto-tsuruoka-2016-adaptive}
Kazuma Hashimoto and Yoshimasa Tsuruoka. 2016.
\newblock \href {https://doi.org/10.18653/v1/P16-1020} {Adaptive joint learning
  of compositional and non-compositional phrase embeddings}.
\newblock In \emph{Proceedings of the 54th Annual Meeting of the Association
  for Computational Linguistics (Volume 1: Long Papers)}, pages 205--215,
  Berlin, Germany. Association for Computational Linguistics.

\bibitem[{Houlsby et~al.(2019)Houlsby, Giurgiu, Jastrzebski, Morrone,
  De~Laroussilhe, Gesmundo, Attariyan, and Gelly}]{pmlr-v97-houlsby19a}
Neil Houlsby, Andrei Giurgiu, Stanislaw Jastrzebski, Bruna Morrone, Quentin
  De~Laroussilhe, Andrea Gesmundo, Mona Attariyan, and Sylvain Gelly. 2019.
\newblock \href {https://proceedings.mlr.press/v97/houlsby19a.html}
  {Parameter-efficient transfer learning for {NLP}}.
\newblock In \emph{Proceedings of the 36th International Conference on Machine
  Learning}, volume~97 of \emph{Proceedings of Machine Learning Research},
  pages 2790--2799. PMLR.

\bibitem[{Hupkes and Zuidema(2018)}]{ijcai2018-796}
Dieuwke Hupkes and Willem Zuidema. 2018.
\newblock \href {https://doi.org/10.24963/ijcai.2018/796} {Visualisation and
  'diagnostic classifiers' reveal how recurrent and recursive neural networks
  process hierarchical structure (extended abstract)}.
\newblock In \emph{Proceedings of the Twenty-Seventh International Joint
  Conference on Artificial Intelligence, {IJCAI-18}}, pages 5617--5621.
  International Joint Conferences on Artificial Intelligence Organization.

\bibitem[{Jana et~al.(2019)Jana, Puzyrev, Panchenko, Goyal, Biemann, and
  Mukherjee}]{jana-etal-2019-compositionality}
Abhik Jana, Dima Puzyrev, Alexander Panchenko, Pawan Goyal, Chris Biemann, and
  Animesh Mukherjee. 2019.
\newblock \href {https://doi.org/10.18653/v1/P19-1316} {On the compositionality
  prediction of noun phrases using poincar{\'e} embeddings}.
\newblock In \emph{Proceedings of the 57th Annual Meeting of the Association
  for Computational Linguistics}, pages 3263--3274, Florence, Italy.
  Association for Computational Linguistics.

\bibitem[{Jhamtani et~al.(2021)Jhamtani, Gangal, Hovy, and
  Berg-Kirkpatrick}]{jhamtani-etal-2021-investigating}
Harsh Jhamtani, Varun Gangal, Eduard Hovy, and Taylor Berg-Kirkpatrick. 2021.
\newblock \href {https://doi.org/10.18653/v1/2021.emnlp-main.592}
  {Investigating robustness of dialog models to popular figurative language
  constructs}.
\newblock In \emph{Proceedings of the 2021 Conference on Empirical Methods in
  Natural Language Processing}, pages 7476--7485, Online and Punta Cana,
  Dominican Republic. Association for Computational Linguistics.

\bibitem[{Lewis et~al.(2020)Lewis, Liu, Goyal, Ghazvininejad, Mohamed, Levy,
  Stoyanov, and Zettlemoyer}]{lewis-etal-2020-bart}
Mike Lewis, Yinhan Liu, Naman Goyal, Marjan Ghazvininejad, Abdelrahman Mohamed,
  Omer Levy, Veselin Stoyanov, and Luke Zettlemoyer. 2020.
\newblock \href {https://doi.org/10.18653/v1/2020.acl-main.703} {{BART}:
  Denoising sequence-to-sequence pre-training for natural language generation,
  translation, and comprehension}.
\newblock In \emph{Proceedings of the 58th Annual Meeting of the Association
  for Computational Linguistics}, pages 7871--7880, Online. Association for
  Computational Linguistics.

\bibitem[{Li et~al.(2018{\natexlab{a}})Li, Yang, Wang, Wang, Cui, and
  Zhang}]{ijcai2018-576}
Bing Li, Xiaochun Yang, Bin Wang, Wei Wang, Wei Cui, and Xianchao Zhang.
  2018{\natexlab{a}}.
\newblock \href {https://doi.org/10.24963/ijcai.2018/576} {An adaptive
  hierarchical compositional model for phrase embedding}.
\newblock In \emph{Proceedings of the Twenty-Seventh International Joint
  Conference on Artificial Intelligence, {IJCAI-18}}, pages 4144--4151.
  International Joint Conferences on Artificial Intelligence Organization.

\bibitem[{Li et~al.(2018{\natexlab{b}})Li, Lu, Xiong, and Long}]{li2018phrase}
Minglei Li, Qin Lu, Dan Xiong, and Yunfei Long. 2018{\natexlab{b}}.
\newblock Phrase embedding learning based on external and internal context with
  compositionality constraint.
\newblock \emph{Knowledge-Based Systems}, 152:107--116.

\bibitem[{Lin et~al.(2021)Lin, Ma, Yan, and Chen}]{lin-etal-2021-cate}
Zhenxi Lin, Qianli Ma, Jiangyue Yan, and Jieyu Chen. 2021.
\newblock \href {https://doi.org/10.18653/v1/2021.emnlp-main.316} {{CATE}: A
  contrastive pre-trained model for metaphor detection with semi-supervised
  learning}.
\newblock In \emph{Proceedings of the 2021 Conference on Empirical Methods in
  Natural Language Processing}, pages 3888--3898, Online and Punta Cana,
  Dominican Republic. Association for Computational Linguistics.

\bibitem[{Liu(2019)}]{liu2019toward}
Changsheng Liu. 2019.
\newblock \emph{Toward Robust and Efficient Interpretations of Idiomatic
  Expressions in Context}.
\newblock Ph.D. thesis, University of Pittsburgh.

\bibitem[{Liu and Hwa(2017)}]{liu2017representations}
Changsheng Liu and Rebecca Hwa. 2017.
\newblock Representations of context in recognizing the figurative and literal
  usages of idioms.
\newblock In \emph{Proceedings of the AAAI Conference on Artificial
  Intelligence}, volume~31.

\bibitem[{Liu and Hwa(2019)}]{liu2019generalized}
Changsheng Liu and Rebecca Hwa. 2019.
\newblock A generalized idiom usage recognition model based on semantic
  compatibility.
\newblock In \emph{Proceedings of the AAAI Conference on Artificial
  Intelligence}, volume~33, pages 6738--6745.

\bibitem[{Liu et~al.(2017)Liu, Qian, Qiu, and Huang}]{liu2017idiom}
Pengfei Liu, Kaiyu Qian, Xipeng Qiu, and Xuan-Jing Huang. 2017.
\newblock Idiom-aware compositional distributed semantics.
\newblock In \emph{Proceedings of the 2017 conference on empirical methods in
  natural language processing}, pages 1204--1213.

\bibitem[{Liu et~al.(2021)Liu, Yuan, Fu, Jiang, Hayashi, and
  Neubig}]{Liu2021PretrainPA}
Pengfei Liu, Weizhe Yuan, Jinlan Fu, Zhengbao Jiang, Hiroaki Hayashi, and
  Graham Neubig. 2021.
\newblock Pre-train, prompt, and predict: A systematic survey of prompting
  methods in natural language processing.
\newblock \emph{ArXiv}, abs/2107.13586.

\bibitem[{Moon et~al.(1998)}]{moon1998fixed}
Rosamund Moon et~al. 1998.
\newblock \emph{Fixed {E}xpressions and {I}dioms in {E}nglish: A
  {C}orpus-{B}ased {A}pproach}.
\newblock Oxford University Press.

\bibitem[{Nedumpozhimana et~al.(2022)Nedumpozhimana, Klubička, and
  Kelleher}]{10.3389/frai.2022.813967}
Vasudevan Nedumpozhimana, Filip Klubička, and John~D. Kelleher. 2022.
\newblock \href {https://doi.org/10.3389/frai.2022.813967} {Shapley idioms:
  Analysing bert sentence embeddings for general idiom token identification}.
\newblock \emph{Frontiers in Artificial Intelligence}, 5.

\bibitem[{Pedregosa et~al.(2011)Pedregosa, Varoquaux, Gramfort, Michel,
  Thirion, Grisel, Blondel, Prettenhofer, Weiss, Dubourg, Vanderplas, Passos,
  Cournapeau, Brucher, Perrot, and Duchesnay}]{scikit-learn}
F.~Pedregosa, G.~Varoquaux, A.~Gramfort, V.~Michel, B.~Thirion, O.~Grisel,
  M.~Blondel, P.~Prettenhofer, R.~Weiss, V.~Dubourg, J.~Vanderplas, A.~Passos,
  D.~Cournapeau, M.~Brucher, M.~Perrot, and E.~Duchesnay. 2011.
\newblock Scikit-learn: Machine learning in {P}ython.
\newblock \emph{Journal of Machine Learning Research}, 12:2825--2830.

\bibitem[{Peng and Feldman(2016)}]{DBLP:conf/simbig/PengF16a}
Jing Peng and Anna Feldman. 2016.
\newblock \href {https://doi.org/10.1007/978-3-319-55209-5\_2} {Automatic idiom
  recognition with word embeddings}.
\newblock In \emph{Information Management and Big Data - Second Annual
  International Symposium, SIMBig 2015, Cusco, Peru, September 2-4, 2015, and
  Third Annual International Symposium, SIMBig 2016, Cusco, Peru, September
  1-3, 2016, Revised Selected Papers}, volume 656 of \emph{Communications in
  Computer and Information Science}, pages 17--29. Springer.

\bibitem[{Peng et~al.(2014)Peng, Feldman, and Vylomova}]{peng2018classifying}
Jing Peng, Anna Feldman, and Ekaterina Vylomova. 2014.
\newblock \href {"https://aclanthology.org/D14-1216} {Classifying idiomatic and
  literal expressions using topic models and intensity of emotions}.
\newblock In \emph{Proceedings of the 2014 Conference on Empirical Methods in
  Natural Language Processing ({EMNLP})}, pages 2019--2027. Association for
  Computational Linguistics.

\bibitem[{Pfeiffer et~al.(2020{\natexlab{a}})Pfeiffer, R{\"u}ckl{\'e}, Poth,
  Kamath, Vuli{\'c}, Ruder, Cho, and Gurevych}]{pfeiffer-etal-2020-adapterhub}
Jonas Pfeiffer, Andreas R{\"u}ckl{\'e}, Clifton Poth, Aishwarya Kamath, Ivan
  Vuli{\'c}, Sebastian Ruder, Kyunghyun Cho, and Iryna Gurevych.
  2020{\natexlab{a}}.
\newblock \href {https://doi.org/10.18653/v1/2020.emnlp-demos.7}
  {{A}dapter{H}ub: A framework for adapting transformers}.
\newblock In \emph{Proceedings of the 2020 Conference on Empirical Methods in
  Natural Language Processing: System Demonstrations}, pages 46--54, Online.
  Association for Computational Linguistics.

\bibitem[{Pfeiffer et~al.(2020{\natexlab{b}})Pfeiffer, R\"uckl\'{e}, Poth,
  Kamath, Vuli\'{c}, Ruder, Cho, and Gurevych}]{pfeiffer2020AdapterHub}
Jonas Pfeiffer, Andreas R\"uckl\'{e}, Clifton Poth, Aishwarya Kamath, Ivan
  Vuli\'{c}, Sebastian Ruder, Kyunghyun Cho, and Iryna Gurevych.
  2020{\natexlab{b}}.
\newblock \href {https://www.aclweb.org/anthology/2020.emnlp-demos.7}
  {Adapterhub: A framework for adapting transformers}.
\newblock In \emph{Proceedings of the 2020 Conference on Empirical Methods in
  Natural Language Processing (EMNLP 2020): Systems Demonstrations}, pages
  46--54, Online. Association for Computational Linguistics.

\bibitem[{Pfeiffer et~al.(2020{\natexlab{c}})Pfeiffer, Vuli{\'c}, Gurevych, and
  Ruder}]{pfeiffer-etal-2020-mad}
Jonas Pfeiffer, Ivan Vuli{\'c}, Iryna Gurevych, and Sebastian Ruder.
  2020{\natexlab{c}}.
\newblock \href {https://doi.org/10.18653/v1/2020.emnlp-main.617} {{MAD-X}:
  {A}n {A}dapter-{B}ased {F}ramework for {M}ulti-{T}ask {C}ross-{L}ingual
  {T}ransfer}.
\newblock In \emph{Proceedings of the 2020 Conference on Empirical Methods in
  Natural Language Processing (EMNLP)}, pages 7654--7673, Online. Association
  for Computational Linguistics.

\bibitem[{Radford et~al.(2021)Radford, Kim, Hallacy, Ramesh, Goh, Agarwal,
  Sastry, Askell, Mishkin, Clark, Krueger, and
  Sutskever}]{Radford2021LearningTV}
Alec Radford, Jong~Wook Kim, Chris Hallacy, Aditya Ramesh, Gabriel Goh,
  Sandhini Agarwal, Girish Sastry, Amanda Askell, Pamela Mishkin, Jack Clark,
  Gretchen Krueger, and Ilya Sutskever. 2021.
\newblock Learning transferable visual models from natural language
  supervision.
\newblock In \emph{ICML}.

\bibitem[{Rajani et~al.(2014)Rajani, Salinas, and Mooney}]{rajani2014using}
Nazneen~Fatema Rajani, Edaena Salinas, and Raymond Mooney. 2014.
\newblock Using abstract context to detect figurative language.

\bibitem[{Ravichander et~al.(2021)Ravichander, Belinkov, and
  Hovy}]{ravichander-etal-2021-probing}
Abhilasha Ravichander, Yonatan Belinkov, and Eduard Hovy. 2021.
\newblock \href {https://doi.org/10.18653/v1/2021.eacl-main.295} {Probing the
  probing paradigm: Does probing accuracy entail task relevance?}
\newblock In \emph{Proceedings of the 16th Conference of the European Chapter
  of the Association for Computational Linguistics: Main Volume}, pages
  3363--3377, Online. Association for Computational Linguistics.

\bibitem[{Rebuffi et~al.(2017)Rebuffi, Bilen, and Vedaldi}]{NIPS2017_e7b24b11}
Sylvestre-Alvise Rebuffi, Hakan Bilen, and Andrea Vedaldi. 2017.
\newblock \href
  {https://proceedings.neurips.cc/paper/2017/file/e7b24b112a44fdd9ee93bdf998c6ca0e-Paper.pdf}
  {Learning multiple visual domains with residual adapters}.
\newblock In \emph{Advances in Neural Information Processing Systems},
  volume~30. Curran Associates, Inc.

\bibitem[{Rebuffi et~al.(2018)Rebuffi, Vedaldi, and Bilen}]{8578945}
Sylvestre-Alvise Rebuffi, Andrea Vedaldi, and Hakan Bilen. 2018.
\newblock \href {https://doi.org/10.1109/CVPR.2018.00847} {Efficient
  parametrization of multi-domain deep neural networks}.
\newblock In \emph{2018 IEEE/CVF Conference on Computer Vision and Pattern
  Recognition}, pages 8119--8127.

\bibitem[{Reddy et~al.(2011)Reddy, McCarthy, and
  Manandhar}]{reddy2011empirical}
Siva Reddy, Diana McCarthy, and Suresh Manandhar. 2011.
\newblock \href {https://aclanthology.org/I11-1024} {An empirical study on
  compositionality in compound nouns}.
\newblock In \emph{Proceedings of 5th International Joint Conference on Natural
  Language Processing}, pages 210--218. Asian Federation of Natural Language
  Processing.

\bibitem[{Reimers and Gurevych(2019)}]{reimers-2019-sentence-bert}
Nils Reimers and Iryna Gurevych. 2019.
\newblock \href {http://arxiv.org/abs/1908.10084} {Sentence-bert: Sentence
  embeddings using siamese bert-networks}.
\newblock In \emph{Proceedings of the 2019 Conference on Empirical Methods in
  Natural Language Processing}. Association for Computational Linguistics.

\bibitem[{Sag et~al.(2002)Sag, Baldwin, Bond, Copestake, and
  Flickinger}]{sag2002multiword}
Ivan~A Sag, Timothy Baldwin, Francis Bond, Ann Copestake, and Dan Flickinger.
  2002.
\newblock Multiword expressions: A pain in the neck for {NLP}.
\newblock In \emph{International conference on intelligent text processing and
  computational linguistics}, pages 1--15. Springer.

\bibitem[{Salton et~al.(2014)Salton, Ross, and Kelleher}]{salton2014empirical}
Giancarlo Salton, Robert Ross, and John Kelleher. 2014.
\newblock \href {https://aclanthology.org/W14-1007} {An empirical study of the
  impact of idioms on phrase based statistical machine translation of {E}nglish
  to {B}razilian-{P}ortuguese}.
\newblock In \emph{Proceedings of the 3rd Workshop on Hybrid Approaches to
  Machine Translation ({H}y{T}ra)}, pages 36--41. Association for Computational
  Linguistics".

\bibitem[{Salton et~al.(2016)Salton, Ross, and Kelleher}]{salton2016idiom}
Giancarlo Salton, Robert Ross, and John Kelleher. 2016.
\newblock Idiom token classification using sentential distributed semantics.
\newblock In \emph{Proceedings of the 54th Annual Meeting of the Association
  for Computational Linguistics (Volume 1: Long Papers)}, pages 194--204.

\bibitem[{Schnabel et~al.(2015)Schnabel, Labutov, Mimno, and
  Joachims}]{schnabel-etal-2015-evaluation}
Tobias Schnabel, Igor Labutov, David Mimno, and Thorsten Joachims. 2015.
\newblock \href {https://doi.org/10.18653/v1/D15-1036} {Evaluation methods for
  unsupervised word embeddings}.
\newblock In \emph{Proceedings of the 2015 Conference on Empirical Methods in
  Natural Language Processing}, pages 298--307, Lisbon, Portugal. Association
  for Computational Linguistics.

\bibitem[{Schneider and Smith(2015)}]{schneider-smith-2015-corpus}
Nathan Schneider and Noah~A. Smith. 2015.
\newblock \href {https://doi.org/10.3115/v1/N15-1177} {A corpus and model
  integrating multiword expressions and supersenses}.
\newblock In \emph{Proceedings of the 2015 Conference of the North {A}merican
  Chapter of the Association for Computational Linguistics: Human Language
  Technologies}, pages 1537--1547, Denver, Colorado. Association for
  Computational Linguistics.

\bibitem[{Sennrich et~al.(2016)Sennrich, Haddow, and
  Birch}]{sennrich-etal-2016-neural}
Rico Sennrich, Barry Haddow, and Alexandra Birch. 2016.
\newblock \href {https://doi.org/10.18653/v1/P16-1162} {Neural machine
  translation of rare words with subword units}.
\newblock In \emph{Proceedings of the 54th Annual Meeting of the Association
  for Computational Linguistics (Volume 1: Long Papers)}, pages 1715--1725,
  Berlin, Germany. Association for Computational Linguistics.

\bibitem[{Shutova et~al.(2010)Shutova, Sun, and Korhonen}]{shutova2010metaphor}
Ekaterina Shutova, Lin Sun, and Anna Korhonen. 2010.
\newblock Metaphor identification using verb and noun clustering.
\newblock In \emph{Proceedings of the 23rd International Conference on
  Computational Linguistics (Coling 2010)}, pages 1002--1010.

\bibitem[{Song et~al.(2020)Song, Tan, Qin, Lu, and
  Liu}]{DBLP:conf/nips/Song0QLL20}
Kaitao Song, Xu~Tan, Tao Qin, Jianfeng Lu, and Tie{-}Yan Liu. 2020.
\newblock \href
  {https://proceedings.neurips.cc/paper/2020/hash/c3a690be93aa602ee2dc0ccab5b7b67e-Abstract.html}
  {Mpnet: Masked and permuted pre-training for language understanding}.
\newblock In \emph{Advances in Neural Information Processing Systems 33: Annual
  Conference on Neural Information Processing Systems 2020, NeurIPS 2020,
  December 6-12, 2020, virtual}.

\bibitem[{Tabossi et~al.(2008)Tabossi, Fanari, and
  Wolf}]{tabossi2008processing}
Patrizia Tabossi, Rachele Fanari, and Kinou Wolf. 2008.
\newblock Processing idiomatic expressions: Effects of semantic
  compositionality.
\newblock \emph{Journal of Experimental Psychology: Learning, Memory, and
  Cognition}, 34(2):313.

\bibitem[{Tabossi et~al.(2009)Tabossi, Fanari, and Wolf}]{tabossi2009idioms}
Patrizia Tabossi, Rachele Fanari, and Kinou Wolf. 2009.
\newblock Why are idioms recognized fast?
\newblock \emph{Memory \& Cognition}, 37(4):529--540.

\bibitem[{Taslimipoor et~al.(2018)Taslimipoor, Rohanian, Mitkov, and
  Fazly}]{taslimipoor2018identification}
Shiva Taslimipoor, Omid Rohanian, Ruslan Mitkov, and Afsaneh Fazly. 2018.
\newblock Identification of multiword expressions: A fresh look at modelling
  and evaluation.
\newblock In \emph{Multiword expressions at length and in depth: Extended
  papers from the MWE 2017 workshop}, volume~2, page 299. Language Science
  Press.

\bibitem[{Tayyar~Madabushi et~al.(2021)Tayyar~Madabushi, Gow-Smith, Scarton,
  and
  Villavicencio}]{tayyar-madabushi-etal-2021-astitchinlanguagemodels-dataset}
Harish Tayyar~Madabushi, Edward Gow-Smith, Carolina Scarton, and Aline
  Villavicencio. 2021.
\newblock \href {https://doi.org/10.18653/v1/2021.findings-emnlp.294}
  {{AS}titch{I}n{L}anguage{M}odels: Dataset and methods for the exploration of
  idiomaticity in pre-trained language models}.
\newblock In \emph{Findings of the Association for Computational Linguistics:
  EMNLP 2021}, pages 3464--3477, Punta Cana, Dominican Republic. Association
  for Computational Linguistics.

\bibitem[{Wang et~al.(2019)Wang, Wang, Chen, Wang, and
  Kuo}]{wang_wang_chen_wang_kuo_2019}
Bin Wang, Angela Wang, Fenxiao Chen, Yuncheng Wang, and C.-C.~Jay Kuo. 2019.
\newblock \href {https://doi.org/10.1017/ATSIP.2019.12} {Evaluating word
  embedding models: methods and experimental results}.
\newblock \emph{APSIPA Transactions on Signal and Information Processing},
  8:e19.

\bibitem[{Warstadt et~al.(2019)Warstadt, Cao, Grosu, Peng, Blix, Nie, Alsop,
  Bordia, Liu, Parrish, Wang, Phang, Mohananey, Htut, Jeretic, and
  Bowman}]{warstadt-etal-2019-investigating}
Alex Warstadt, Yu~Cao, Ioana Grosu, Wei Peng, Hagen Blix, Yining Nie, Anna
  Alsop, Shikha Bordia, Haokun Liu, Alicia Parrish, Sheng-Fu Wang, Jason Phang,
  Anhad Mohananey, Phu~Mon Htut, Paloma Jeretic, and Samuel~R. Bowman. 2019.
\newblock \href {https://doi.org/10.18653/v1/D19-1286} {Investigating
  {BERT}{'}s knowledge of language: Five analysis methods with {NPI}s}.
\newblock In \emph{Proceedings of the 2019 Conference on Empirical Methods in
  Natural Language Processing and the 9th International Joint Conference on
  Natural Language Processing (EMNLP-IJCNLP)}, pages 2877--2887, Hong Kong,
  China. Association for Computational Linguistics.

\bibitem[{Westerst{\aa}hl(2002)}]{westerstaahl2002compositionality}
Dag Westerst{\aa}hl. 2002.
\newblock On the compositionality of idioms.
\newblock \emph{Proceedings of LLC8. CSLI Publications}.

\bibitem[{Wolf et~al.(2020)Wolf, Debut, Sanh, Chaumond, Delangue, Moi, Cistac,
  Rault, Louf, Funtowicz, Davison, Shleifer, von Platen, Ma, Jernite, Plu, Xu,
  Scao, Gugger, Drame, Lhoest, and Rush}]{wolf-etal-2020-transformers}
Thomas Wolf, Lysandre Debut, Victor Sanh, Julien Chaumond, Clement Delangue,
  Anthony Moi, Pierric Cistac, Tim Rault, Rémi Louf, Morgan Funtowicz, Joe
  Davison, Sam Shleifer, Patrick von Platen, Clara Ma, Yacine Jernite, Julien
  Plu, Canwen Xu, Teven~Le Scao, Sylvain Gugger, Mariama Drame, Quentin Lhoest,
  and Alexander~M. Rush. 2020.
\newblock \href {https://www.aclweb.org/anthology/2020.emnlp-demos.6}
  {Transformers: State-of-the-art natural language processing}.
\newblock In \emph{Proceedings of the 2020 Conference on Empirical Methods in
  Natural Language Processing: System Demonstrations}, pages 38--45, Online.
  Association for Computational Linguistics.

\bibitem[{Wu et~al.(2016)Wu, Schuster, Chen, Le, Norouzi, Macherey, Krikun,
  Cao, Gao, Macherey, Klingner, Shah, Johnson, Liu, Kaiser, Gouws, Kato, Kudo,
  Kazawa, Stevens, Kurian, Patil, Wang, Young, Smith, Riesa, Rudnick, Vinyals,
  Corrado, Hughes, and Dean}]{DBLP:journals/corr/WuSCLNMKCGMKSJL16}
Yonghui Wu, Mike Schuster, Zhifeng Chen, Quoc~V. Le, Mohammad Norouzi, Wolfgang
  Macherey, Maxim Krikun, Yuan Cao, Qin Gao, Klaus Macherey, Jeff Klingner,
  Apurva Shah, Melvin Johnson, Xiaobing Liu, Lukasz Kaiser, Stephan Gouws,
  Yoshikiyo Kato, Taku Kudo, Hideto Kazawa, Keith Stevens, George Kurian,
  Nishant Patil, Wei Wang, Cliff Young, Jason Smith, Jason Riesa, Alex Rudnick,
  Oriol Vinyals, Greg Corrado, Macduff Hughes, and Jeffrey Dean. 2016.
\newblock \href {http://arxiv.org/abs/1609.08144} {Google's neural machine
  translation system: Bridging the gap between human and machine translation}.
\newblock \emph{CoRR}, abs/1609.08144.

\bibitem[{Yu et~al.(2021)Yu, Zhu, Fang, Yu, Wang, Xu, Zeng, and
  Jiang}]{Yu2021DictBERTEL}
W.~Yu, Chenguang Zhu, Yuwei Fang, Donghan Yu, Shuohang Wang, Yichong Xu,
  Michael Zeng, and Meng Jiang. 2021.
\newblock Dict-bert: Enhancing language model pre-training with dictionary.
\newblock \emph{ArXiv}, abs/2110.06490.

\bibitem[{Zeng and Bhat(2021)}]{zeng-bhat-2021-idiomatic}
Ziheng Zeng and Suma Bhat. 2021.
\newblock \href {https://doi.org/10.1162/tacl_a_00442} {Idiomatic expression
  identification using semantic compatibility}.
\newblock \emph{Transactions of the Association for Computational Linguistics},
  9:1546--1562.

\bibitem[{Zhou et~al.(2021)Zhou, Zeng, Gong, and Bhat}]{zhou2021idiomatic}
Jianing Zhou, Ziheng Zeng, Hongyu Gong, and Suma Bhat. 2021.
\newblock \href {http://arxiv.org/abs/2112.08592} {Idiomatic expression
  paraphrasing without strong supervision}.

\bibitem[{Škvorc et~al.(2022)Škvorc, Gantar, and
  Robnik-Šikonja}]{SKVORC2022107606}
Tadej Škvorc, Polona Gantar, and Marko Robnik-Šikonja. 2022.
\newblock \href {https://doi.org/https://doi.org/10.1016/j.knosys.2021.107606}
  {Mice: Mining idioms with contextual embeddings}.
\newblock \emph{Knowledge-Based Systems}, 235:107606.

\end{thebibliography}
\bibliographystyle{acl_natbib}
\end{document}